\documentclass[conference,a4paper]{IEEEtran}
\IEEEoverridecommandlockouts

\usepackage{cite}
\usepackage{amsmath,amssymb,amsfonts}
\usepackage{algorithmic}
\usepackage{graphicx}
\usepackage{textcomp}
\usepackage{xcolor}
\usepackage[table]{xcolor}
\usepackage{booktabs}
\usepackage{multirow}
\usepackage{soul}   
\usepackage{enumitem}  
\usepackage{tikz}
\usepackage[margin=0.8in]{geometry}
\def\BibTeX{{\rm B\kern-.05em{\sc i\kern-.025em b}\kern-.08em
    T\kern-.1667em\lower.7ex\hbox{E}\kern-.125emX}}

\definecolor{gray10}{gray}{0.9} 
\definecolor{gray20}{gray}{0.8}
\definecolor{gray30}{gray}{0.7}

\newcommand{\shadecell}[1]{%
  \ifdim #1pt > 0.95pt \cellcolor{gray30}#1%
  \else\ifdim #1pt > 0.90pt \cellcolor{gray20}#1%
  \else\ifdim #1pt > 0.80pt \cellcolor{gray10}#1%
  \else #1%
  \fi\fi\fi}

\newcommand\copyrighttext{%
  \footnotesize \textcopyright \the\year{} IEEE. Personal use of this material is permitted. Permission from IEEE must be obtained for all other uses, including reprinting/republishing this material for advertising or promotional purposes, collecting new collected works for resale or redistribution to servers or lists, or reuse of any copyrighted component of this work in other works.}

\newcommand\copyrightnotice{%
\begin{tikzpicture}[remember picture,overlay]
\node[anchor=south,yshift=10pt] at (current page.south) {\fbox{\parbox{\dimexpr0.9\textwidth-\fboxsep-\fboxrule\relax}{\copyrighttext}}};
\end{tikzpicture}%
}

\begin{document}

\title{Investigating Adversarial Robustness against Preprocessing used in Blackbox Face Recognition}

\author{\IEEEauthorblockN{Roland Croft}
\IEEEauthorblockA{\textit{Swordfish Computing} \\
Adelaide, Australia \\
\fontsize{9}{10}\selectfont
roland.croft@swordfish.com.au}
\and
\IEEEauthorblockN{Brian Du}
\IEEEauthorblockA{\textit{Swordfish Computing} \\
Adelaide, Australia \\
\fontsize{9}{10}\selectfont
brian.du@swordfish.com.au}
\and
\IEEEauthorblockN{Darcy Joseph}
\IEEEauthorblockA{\textit{Swordfish Computing} \\
Adelaide, Australia \\
\fontsize{9}{10}\selectfont
darcy.joseph@swordfish.com.au}
\and
\IEEEauthorblockN{Sharath Kumar}
\IEEEauthorblockA{\textit{Swordfish Computing} \\
Adelaide, Australia \\
\fontsize{9}{10}\selectfont
sharath.kumar@swordfish.com.au}
}

\maketitle

\copyrightnotice

\begin{figure*}[!htb]
  \centering
  \includegraphics[width=0.9\textwidth]{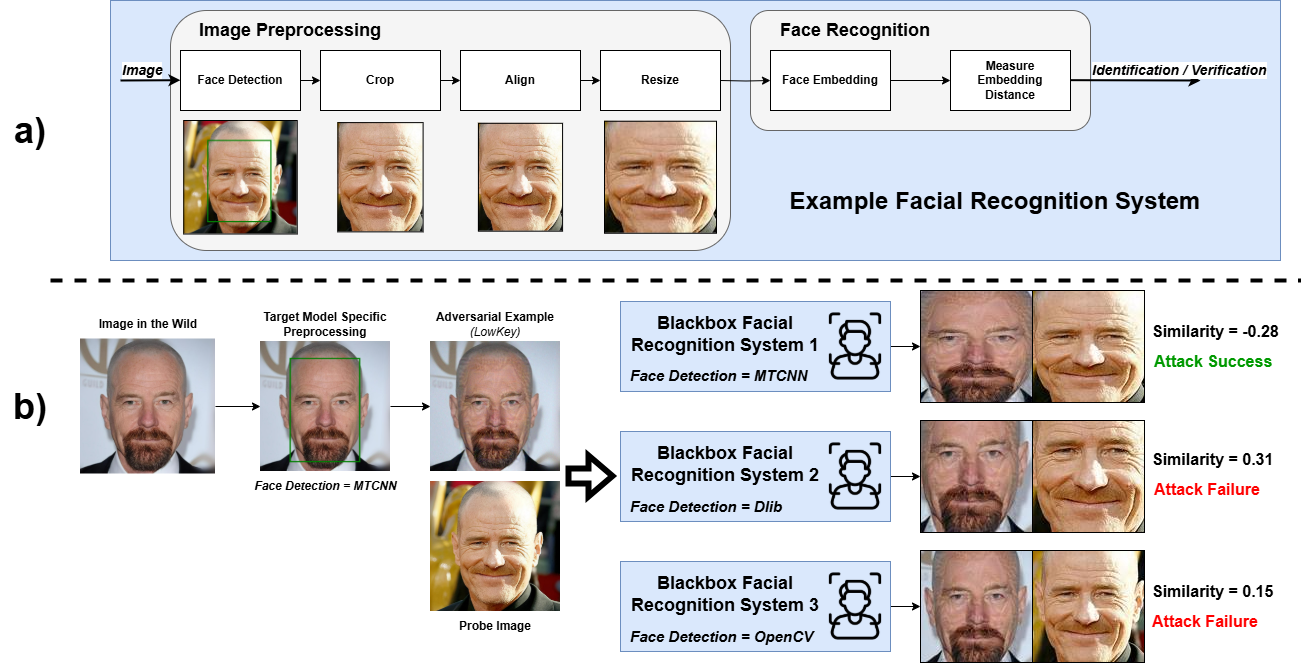}
  \caption{a) The structure of a face recognition system \cite{kortli2020face}. b) Example effects on adversarial FR attacks against different preprocessing methods used in blackbox FR systems, assuming a verification threshold of $> 0.15$.}
  \label{fig:motivation}
\end{figure*}

\begin{abstract}
Face Recognition (FR) models have been shown to be vulnerable to adversarial examples that subtly alter benign facial images, exposing blind spots in these systems, as well as protecting user privacy. End-to-end FR systems first obtain preprocessed faces from diverse facial imagery prior to computing the similarity of the deep feature embeddings. Whilst face preprocessing is a critical component of FR systems, and hence adversarial attacks against them, we observe that this preprocessing is often overlooked in blackbox settings. Our study seeks to investigate the transferability of several out-of-the-box state-of-the-art adversarial attacks against FR when applied against different preprocessing techniques used in a blackbox setting. We observe that the choice of face detection model can degrade the attack success rate by up to 78\%, whereas choice of interpolation method during downsampling has relatively minimal impacts. Furthermore, we find that the requirement for facial preprocessing even degrades attack strength in a whitebox setting, due to the unintended interaction of produced noise vectors against face detection models. Based on these findings, we propose a preprocessing-invariant method using input transformations that improves the transferability of the studied attacks by up to 27\%. Our findings highlight the importance of preprocessing in FR systems, and the need for its consideration towards improving the adversarial generalisation of facial adversarial examples. 
\end{abstract}


\section{Introduction}

Face recognition (FR) systems have gained significant interest due to their various useful applications, such as video surveillance, building access control, and personal identification. The capabilities of these systems have been advanced through the application of Deep Learning (DL)-based feature extraction, to enable FR algorithms to strongly interpret facial features \cite{kortli2020face}. However, these advancements, alongside the growth in widespread use of image acquisition technologies have raised serious privacy concerns for individuals and their personal online imagery \cite{cao2024face}. Consequently, significant effort has been made towards methods for face de-identification or privacy protection \cite{shan2020fawkes, sun2024diffam}.

Many studies have leveraged adversarial examples to achieve facial privacy protection \cite{vakhshiteh2021adversarial, xu2022adversarial, cao2024face, wen2024face}, which overlay adversarial perturbations on an original image to exploit vulnerabilities in FR models. These methods have been shown to effectively prevent FR systems from making correct predictions, whilst making minimal alterations to the original image \cite{shan2020fawkes, cherepanova2021lowkey, yang2021towards}, thus preserving perceptual similarity and identity. 

However, a key challenge for adversarial examples is adversarial generalisation \cite{xie2019improving, dong2019evading}; adversarial perturbations commonly have limited transferability to unseen models and applications. Existing works for adversarial attacks against FR typically assume a whitebox setup \cite{cao2024face}. 

Under a blackbox setup, the attacker does not have knowledge of the architecture and setup of the FR system. Whilst some prior works have considered generalisation of adversarial examples for FR in a blackbox setup \cite{yang2020robfr, yang2021towards, zhou2025improving}, the evaluation and scope of this analysis has been limited to transferability against different victim FR models. We observe that facial preprocessing is a critical and ill-considered component of adversarial attacks against FR. An FR system needs to detect, crop, align, resize, and normalise facial images, all before being passed to the actual FR model of interest. These preprocessing steps can be performed using a variety of options, which can lead to significantly different extracted facial features. However, to the best of our knowledge no prior study has considered the impact that these preprocessing steps would impose on a FR system. Figure \ref{fig:motivation}:b provides an example of the impact that different facial preprocessing can have on the FR outputs. 

Hence, we aim to investigate the effect that different image preprocessing methods have on adversarial attacks against FR systems. For the scope of this investigation, we consider two main preprocessing steps: 1) face cropping via different face detection models, and 2) image resizing via different interpolation methods. We conduct extensive experiments to determine the effect of preprocessing on the robustness of adversarial examples against blackbox FR systems. This analysis yields essential insights into adversarial generalisation against FR systems, which we then use to produce more effective image augmentation techniques for improving adversarial robustness.

Our main contributions can be summarised as follows: 
\begin{itemize}
    \item We provide in-depth analysis of the role of image and face preprocessing methods when creating adversarial examples of facial images. 
    \item We demonstrate and measure the impact to performance when using different open-source face detection backends or downsampling interpolation methods against FR adversarial examples to examine degradation of noise vectors against a consistent FR model. 
    \item We propose novel image augmentation techniques for improving the adversarial robustness of adversarial examples against FR systems through preprocessing-related image transformations. 
\end{itemize}

\section{Related Work}
\subsection{Adversarial Machine Learning}

Adversarial attacks introduce subtle but intentional perturbations into benign images, deceiving machine learning systems into producing incorrect predictions \cite{szegedy2013intriguing, goodfellow2014explaining}. An important property of adversarial examples is their transferability; perturbations optimised to attack one model can often deceive other models \cite{szegedy2013intriguing, goodfellow2014explaining, papernot2016transferability}. This property underpins the feasibility of black-box attacks in real-world applications, where the adversary has no access to the architecture or parameters of a target model \cite{papernot2017practical, kurakin2018adversarial}. 

To enhance transferability, Liu et al. \cite{liu2016delving} proposed using ensemble-based approaches, showing that adversarial attacks crafted to attack multiple models generalise better. Subsequent research by Xie et al. \cite{xie2019improving} concluded that iterative methods \cite{kurakin2018adversarial, dong2018boosting} tend to overfit to specific model architectures, resulting in poor transferability. 

To mitigate this overfitting phenomenon, input transformation was introduced into the attack generation process \cite{dong2019evading}. For instance, DI\textsuperscript{2}-FGSM \cite{xie2019improving} applies random resizing and padding to adversarial examples at each iteration, effectively preventing overfitting. Building on this, subsequent research explored other transform types such as noise injection, denoising, contrast equalisation, image compression, geometric distortions to create more robust adversarial examples \cite{dong2019evading, wang2023structure}.

\subsection{Adversarial Attacks against Face Recognition}
Motivated by protecting personal privacy against unauthorised FR, adversarial attacks have been adopted as a countermeasure against FR systems \cite{shan2020fawkes, cao2024face}. In literature, these attacks are categorised into restricted and unrestricted methods \cite{wen2024face, cao2024face, zhou2025improving}. Restricted attacks generate perturbations within a bounded constraint through a noise vector that aims to be visually imperceptible \cite{shan2020fawkes,yang2021towards,yang2020robfr,cherepanova2021lowkey, zhou2023improving, zhou2025improving}. In contrast, unrestricted attacks, do not consider predefined perturbation bounds. These methods include obfuscation-based methods \cite{yuan2022pro}, which apply visually perceptible pixel changes to a face to conceal it, and generative-based methods \cite{xiao2021improving, sun2024diffam}, which modify high-level attributes such as makeup \cite{yin2021adv}, facial expression \cite{jia2022adv}, or lighting \cite{zhang2024adversarial}. However, as these methods are unrestricted, they consequently either degrade the image quality or the perceptual identity, severely inhibiting their practical real-world application. Hence, for the purposes of this study, we focus on restricted adversarial perturbation methods that aim to achieve high perceptual similarity to the original image. 

Among restricted black-box methods, attacks such as LowKey \cite{cherepanova2021lowkey}, TIP-IM \cite{yang2021towards}, BPFA \cite{zhou2023improving}, and DPA \cite{zhou2025improving} incorporate transform-invariant strategies through random transforms such as Gaussian smoothing, affine transformations, and feature augmentation during adversarial example generation, respectively. However, much of the current literature assumes a whitebox setting and operates under ideal conditions \cite{yang2020robfr, yang2021towards, pan2023collaborative}, where datasets are already cropped, aligned and resized, often bypassing the preprocessing of a real system. In practice, most modern FR systems rely on external face detectors to extract the face region before feeding them into embedding networks \cite{kortli2020face}. Variability in the preprocessing stage can significantly impact the effectiveness and transferability of adversarial attacks, which motivates a deeper investigation into how adversarial robustness is influenced by preprocessing.

\section{Methodology}

\subsection{Problem Formulation}

Following the definition of FR by Kortli et al. \cite{kortli2020face}, we consider there to be two main components of an FR system, as depicted in Figure \ref{fig:motivation}:a. 

\begin{enumerate}
    \item \textit{Image Preprocessing:} An FR system begins by standardising images to enable a consistent input to downstream face embedding models. These image inputs to an FR system are commonly referred to as probe images. First, the face and its bounding box are detected using a face detection model. The image is then cropped to that face region and aligned so that the face has a consistent position and structure. The image is then resized to match the input dimensions of the selected face embedding model. 
    
    \item \textit{Face Recognition:} Features are then extracted from the processed facial image using a face embedding model to produce a latent vector representation of the face. Face embedding models, i.e., ArcFace \cite{deng2019arcface}, FaceNet \cite{schroff2015facenet}, etc., are commonly referred to as FR models, due to their prolific use in these systems. Finally, the extracted features of the probe image are compared to the extracted features of a series of known images from a face image gallery database, using distance metrics such as the cosine similarity of the vector embeddings \cite{serengil2024lightface}. There are two main applications of FR \cite{kortli2020face}: \textbf{face verification}, which aims to determine whether two images are of the same person, and \textbf{face identification}, which determines the identity of a probe image. 

\end{enumerate}

Adversarial attacks against FR systems work by generating perturbed images whose feature vectors lie far away from the original image. Maximising the distance of the feature space prevents images from matching other images of the individual. Similarly, adversarial attacks aim to minimise loss of perceptual similarity between the original and perturbed image so that image quality is not degraded. 

Adversarial perturbations are typically generated with respect to a target FR model that is used to guide the attack \cite{yang2020robfr, yang2021towards}. Therefore, these adversarial attacks similarly need to consider image preprocessing whilst generating noise perturbations so that the adversarial example can be produced on non-standardised images \cite{cherepanova2021lowkey}. However, despite prior work conducting significant analysis on the transferability of adversarial attacks against FR models, we observe the image preprocessing component to be ill-considered. 

In this study, we postulate that image preprocessing plays a significant role in adversarial attacks against FR. Hence, we aim to investigate whether inconsistent image preprocessing techniques applied during FR and during generation of adversarial examples has a negative impact on the distance of the produced feature vectors. Any degradation of this feature distance can heavily degrade the success of adversarial examples by decreasing the likelihood of preventing image matches. 

To focus our analysis, we considered adversarial attacks under a whitebox model setup; the adversarial examples are generated using the same FR model as the target FR system. However, we considered image preprocessing under a blackbox setup; the target FR system applies different image preprocessing to the adversarial attack. 

We considered three Research Questions (RQs) to guide our analysis: 
\begin{enumerate}[label=\textbf{RQ\arabic*}:, leftmargin=*, align=left]
  \item \textbf{How does blackbox face detection impact adversarial attack strength against FR systems?}
  \item \textbf{How does blackbox image interpolation impact adversarial attack strength against FR systems?}
  \item \textbf{Is input transformation effective for improving adversarial transferability against different face detection and interpolation methods?}
\end{enumerate}

To limit the scope of this investigation, we perform our experiments over an aligned image dataset, and do not consider this preprocessing step. As we use a consistent FR model that requires a consistent size, we investigate the interpolation process during downsampling. 

\subsection{Considered Attacks}

For our investigation, we considered three state-of-the-art adversarial attacks against facial recognition systems. 1) LowKey \cite{cherepanova2021lowkey}, 2) Momentum Iterative Method (MIM) \cite{dong2018boosting}, and 3) Targeted Identity Protection Iterative Method (TIP-IM) \cite{yang2021towards}. MIM and TIP-IM do not consider image preprocessing in their original papers, as they operated on processed image sets at standardised resolution and crops. Hence, we re-implemented each method to incorporate image preprocessing to enable these attacks to work on real-world images of any shape and size. 

LowKey \cite{cherepanova2021lowkey} uses signed gradient ascent to maximise the feature distance of adversarial examples in an iterative manner. It extends traditional iterative methods \cite{kurakin2018adversarial} by adding ensemble target models, perceptual similarity, and Gaussian blurring to the objective function, to improve transferability. Importantly, LowKey also incorporates face detection, resizing, and alignment as part of the objective function, to enable the attack to have real-world applications to diverse images. We use this as inspiration to alter the other attacks to incorporate preprocessing in a similar way. Formally, the objective function used for LowKey is

\begin{equation}
\resizebox{\columnwidth}{!}{$
\begin{aligned}
  x'_{t+1} = x'_{t} - \alpha \cdot \textrm{sign}(\nabla _{{x}}\mathcal {L}( {x}'_{t}))
\\
\noalign{\vskip 4pt}
  \mathcal {L}( {x}') = 
  \frac{
    \| f(A(x)) - f(A(x')) \|_2^2 +
    \| f(A(x)) - f(A(G(x'))) \|_2^2
  }{
    \| f(A(x)) \|_2
  }
\\
\noalign{\vskip 1pt}
  - \gamma \, \text{LPIPS}(x, x')
\end{aligned}
$}
\label{eq:lowkey}
\end{equation}

where $x$ is the original image, $x'$ is the perturbed image, $t$ denotes the iteration, $f$ denotes the FR model, $\mathcal {L}$ is the loss function, $G$ is the Gaussian smoothing function with fixed parameters, $\gamma$ is the perceptual weighting, and $A$ denotes face detection and extraction followed by resizing and alignment. Equation \ref{eq:lowkey} is altered from the original LowKey implementation to only consider a single target model, due to our whitebox model setup and for consistency with the other considered attacks. 

MIM \cite{dong2018boosting} is an extension of the traditional Fast Gradient Sign Method (FGSM) adversarial attack \cite{goodfellow2014explaining}, which introduces momentum into the iterative process to improve blackbox transferability. Yang et. al \cite{yang2020robfr} demonstrated that MIM has a high success rate in both whitebox and blackbox attacks against face verification. We modified the implementation of the attack by Yang et al. \cite{yang2020robfr} to only require a single image as input, as well as to incorporate image preprocessing, to make the attack more suitable for real-world applications to diverse images. The optimisation function for MIM is formally represented as

\begin{equation}
\begin{aligned}
  x'_{t+1} =  x'_{t} - \alpha \cdot \textrm{sign} (g_{t+1})
\\
\noalign{\vskip 4pt}
  g_{t+1} = \mu \cdot g_t + \nabla_x \mathcal {L}( {x}'_{t})
\\
\noalign{\vskip 4pt}
  \mathcal {L}( {x}') = \frac{(f(A(x)) \cdot f(A(x'))}{\|f(A(x))\| \|f(A(x'))\|}
\end{aligned}
\label{eq:mim} 
\end{equation}

where $g$ is the momentum-based gradient, $\alpha$ is the learning rate, and $\mu$ is the momentum term. 

TIP-IM \cite{yang2021towards} similarly uses an iterative method to maximise the adversarial example feature distance. However, TIP-IM also incorporates an additional set of target images of other human faces, to help guide the noise vector to be more realistic and less perceptually noticeable. Additionally, TIP-IM incorporates image augmentations of rotation and affine transformations at each iteration, to improve blackbox transferability. We adapted the optimisation function from the original TIP-IM implementation as

\begin{equation}
\resizebox{\columnwidth}{!}{$
\begin{aligned}
    x'_{t+1} = x'_{t} - \alpha \cdot \textrm{sign}(\nabla _{{x}}\mathcal {L}( {x}'_{t})) 
\\
    \mathcal {L}(x') =  \frac{1}{N}\sum^{N}_{i=1}{( f(A(x')) - f(A(x^r_{i})) )^2 - ( f(A(x')) - f(A(x)) )^2}
\label{eq:tipim}
\end{aligned}
$}
\end{equation}

where $x^r$ is a real target image. The full implementation details of TIP-IM are provided in the original paper \cite{yang2021towards}. Whilst the original TIP-IM also considers a perceptual loss term via maximum mean discrepancy (MMD) \cite{borgwardt2006integrating}, we did not consider it here as we followed the default attack settings described by Yang et al. \cite{yang2021towards} in which the perceptual weighting is 0. 

\begin{table*}[ht]
\centering
\renewcommand{\arraystretch}{1.2}
\setlength{\tabcolsep}{4pt}
\scriptsize
\resizebox{1.01\textwidth}{!}{%
    \begin{tabular}{|c|c|ccc|ccc|ccc|ccc|ccc|ccc|ccc|}
    \hline
    \multirow{2}{*}{\shortstack{Detection\\Backend}} & \multirow{2}{*}{Attack} & \multicolumn{3}{c|}{MTCNN} & \multicolumn{3}{c|}{OpenCV} & \multicolumn{3}{c|}{Dlib} & \multicolumn{3}{c|}{MediaPipe} & \multicolumn{3}{c|}{YOLO} & \multicolumn{3}{c|}{Centerface} & \multicolumn{3}{c|}{RetinaFace}\\
     &  & I1 & I9 & ASR & I1 & I9 & ASR & I1 & I9 & ASR  & I1 & I9 & ASR & I1 & I9 & ASR & I1 & I9 & ASR  & I1 & I9 & ASR \\
    \hline
    \multirow{3}{*}{MTCNN \cite{zhang2016joint}}
     & LowKey & \textbf{-0.58} & \textbf{-0.25} & \textbf{1.00} & 0.39 & 0.18 & 0.66 & 0.56 & 0.33 & 0.96 & 0.57 & 0.30 & 0.95 & 0.04 & -0.00 & 0.98 & 0.46 & 0.17 & 0.94 & -0.01 & -0.02 & 0.99 \\
     & MIM & \textbf{-0.16} & \textbf{-0.05} & \textbf{1.00} & 0.36 & 0.17 & 0.69 & 0.52 & 0.30 & 0.97 & 0.54 & 0.28 & 0.96 & 0.16 & 0.07 & 0.98 & 0.44 & 0.17 & 0.95 & 0.13 & 0.06 & 0.98 \\
     & TIP-IM & \textbf{-0.01} & \textbf{0.08} & \textbf{0.99} & 0.22 & 0.12 & 0.84 & 0.46 & 0.30 & \textbf{0.99} & 0.48 & 0.29 & 0.98 & 0.11 & 0.10 & \textbf{0.99} & 0.38 & 0.17 & 0.96 & 0.09 & 0.09 & \textbf{0.99} \\
    \hline
    \multirow{3}{*}{OpenCV \cite{bradski2000opencv}}
     & LowKey & 0.35 & 0.14 & 0.94 & \textbf{-0.60} & \textbf{-0.32} & \textbf{1.00} & 0.52 & 0.31 & 0.97 & 0.58 & 0.31 & 0.95 & 0.39 & 0.16 & 0.90 & 0.50 & 0.18 & 0.92 & 0.39 & 0.16 & 0.92 \\
     & MIM & 0.37 & 0.17 & 0.93 & \textbf{-0.18} & \textbf{-0.10} & \textbf{1.00} & 0.50 & 0.29 & 0.98 & 0.55 & 0.28 & 0.97 & 0.40 & 0.18 & 0.90 & 0.48 & 0.18 & 0.94 & 0.40 & 0.18 & 0.90 \\
     & TIP-IM & 0.27 & 0.13 & 0.97 & \textbf{-0.11} & \textbf{-0.04} & \textbf{1.00} & 0.44 & 0.28 & 0.99 & 0.49 & 0.28 & 0.97 & 0.28 & 0.13 & 0.95 & 0.43 & 0.18 & 0.95 & 0.28 & 0.14 & 0.96 \\
    \hline
    \multirow{3}{*}{Dlib \cite{king2009dlib}}
     & LowKey & 0.71 & 0.32 & 0.53 & 0.68 & 0.35 & 0.23 & \textbf{-0.62} & \textbf{-0.31} & \textbf{1.00} & 0.51 & 0.24 & 0.96 & 0.74 & 0.34 & 0.47 & 0.64 & 0.24 & 0.83 & 0.74 & 0.34 & 0.48 \\
     & MIM & 0.62 & 0.29 & 0.64 & 0.58 & 0.30 & 0.32 & \textbf{-0.22} & \textbf{-0.08} & \textbf{1.00} & 0.43 & 0.22 & 0.98 & 0.65 & 0.31 & 0.57 & 0.56 & 0.22 & 0.88 & 0.65 & 0.31 & 0.59 \\
     & TIP-IM & 0.46 & 0.23 & 0.81 & 0.41 & 0.22 & 0.56 & \textbf{0.08} & \textbf{0.17} & \textbf{1.00} & 0.35 & 0.22 & 0.99 & 0.50 & 0.25 & 0.74 & 0.43 & 0.18 & 0.95 & 0.49 & 0.25 & 0.76 \\
    \hline
    \multirow{3}{*}{MediaPipe \cite{lugaresi2019mediapipe}}
     & LowKey & 0.67 & 0.31 & 0.57 & 0.67 & 0.35 & 0.22 & 0.42 & 0.23 & 0.99 & \textbf{-0.60} & \textbf{-0.26} & \textbf{1.00} & 0.68 & 0.32 & 0.53 & 0.62 & 0.24 & 0.85 & 0.68 & 0.32 & 0.54 \\
     & MIM & 0.59 & 0.28 & 0.68 & 0.59 & 0.31 & 0.32 & 0.37 & 0.21 & 0.99 & \textbf{-0.21} & \textbf{-0.07} & \textbf{1.00} & 0.60 & 0.29 & 0.63 & 0.54 & 0.21 & 0.89 & 0.60 & 0.29 & 0.65 \\
     & TIP-IM & 0.46 & 0.24 & 0.81 & 0.43 & 0.23 & 0.54 & 0.35 & 0.25 & \textbf{1.00} & \textbf{0.10} & \textbf{0.19} & \textbf{1.00} & 0.48 & 0.25 & 0.76 & 0.44 & \textbf{0.19} & 0.94 & 0.48 & 0.25 & 0.78 \\
    \hline
    \multirow{3}{*}{YOLO \cite{redmon2016you}}
     & LowKey & -0.00 & -0.02 & \textbf{1.00} & 0.38 & 0.18 & 0.67 & 0.57 & 0.33 & 0.95 & 0.54 & 0.29 & 0.96 & \textbf{-0.56} & \textbf{-0.25} & \textbf{1.00} & 0.49 & 0.18 & 0.94 & -0.23 & -0.11 & \textbf{1.00} \\
     & MIM & 0.14 & 0.06 & 0.99 & 0.36 & 0.17 & 0.70 & 0.53 & 0.31 & 0.97 & 0.51 & 0.27 & 0.97 & \textbf{-0.16} & \textbf{-0.06} & \textbf{1.00} & 0.46 & 0.18 & 0.94 & 0.02 & 0.01 & \textbf{1.00} \\
     & TIP-IM & 0.10 & 0.10 & \textbf{0.99} & 0.22 & 0.12 & 0.84 & 0.48 & 0.32 & 0.98 & 0.49 & 0.30 & 0.97 & \textbf{-0.02} & \textbf{0.07} & \textbf{0.99} & 0.40 & 0.19 & 0.94 & 0.03 & 0.08 & \textbf{0.99} \\
    \hline
    \multirow{3}{*}{Centerface \cite{yuanyuan2019centerface}}
     & LowKey & 0.68 & 0.31 & 0.58 & 0.71 & 0.36 & 0.22 & 0.72 & 0.41 & 0.80 & 0.74 & 0.37 & 0.84 & 0.71 & 0.33 & 0.51 & \textbf{-0.61} & \textbf{-0.20} & \textbf{1.00} & 0.70 & 0.33 & 0.53 \\
     & MIM & 0.60 & 0.28 & 0.67 & 0.63 & 0.32 & 0.31 & 0.65 & 0.37 & 0.88 & 0.66 & 0.35 & 0.89 & 0.63 & 0.30 & 0.59 & \textbf{-0.22} & \textbf{-0.05} & \textbf{1.00} & 0.63 & 0.29 & 0.62 \\
     & TIP-IM & 0.45 & 0.22 & 0.82 & 0.46 & 0.24 & 0.51 & 0.51 & 0.30 & 0.96 & 0.53 & 0.29 & 0.95 & 0.48 & 0.24 & 0.76 & \textbf{-0.00} & \textbf{0.10} & \textbf{1.00} & 0.47 & 0.24 & 0.77 \\
    \hline
    \multirow{3}{*}{RetinaFace \cite{deng2020retinaface}}
     & LowKey & -0.04 & -0.04 & \textbf{1.00} & 0.39 & 0.18 & 0.64 & 0.58 & 0.33 & 0.95 & 0.56 & 0.29 & 0.96 & -0.22 & -0.11 & \textbf{1.00} & 0.49 & 0.17 & 0.93 & \textbf{-0.57} & \textbf{-0.25} & \textbf{1.00} \\
     & MIM & 0.12 & 0.06 & 0.99 & 0.37 & 0.18 & 0.68 & 0.53 & 0.31 & 0.97 & 0.52 & 0.28 & 0.96 & 0.02 & 0.02 & \textbf{1.00} & 0.46 & 0.18 & 0.94 & \textbf{-0.16} & \textbf{-0.05} & \textbf{1.00} \\
     & TIP-IM & 0.06 & 0.05 & \textbf{1.00} & 0.23 & 0.11 & 0.86 & 0.46 & 0.29 & 0.99 & 0.45 & 0.27 & 0.98 & 0.01 & 0.04 & \textbf{1.00} & 0.39 & 0.16 & 0.96 & \textbf{-0.04} & \textbf{0.03} & \textbf{1.00} \\
    \hline
    \end{tabular}
}
\vspace{2pt}
\caption{I1/I9 image similarity scores and Attack Success Rate (ASR) for different attacks with different detection backends. Rows indicate the face detection algorithm used by the adversarial attack, whereas columns indicate the face detection algorithm used by the FR system. Lower is better for I1/I9 whereas higher is better for ASR. \textbf{Bold values} indicate the best row-wise performance for each individual metric.}
\label{tab:detection-method-comparison}
\end{table*}

\subsection{Preprocessing Invariant Attack Method}
\label{sec:preprocessing_invariant_method}

Finally, we considered how to improve the transferability of adversarial examples against unknown preprocessing steps in a blackbox setup. Prior work has shown that input diversity is effective for improving adversarial generalisation \cite{xie2019improving, dong2018boosting}. Random input transformations can be applied to adversarial images at each iteration to help prevent the adversarial perturbation from overfitting to the target whitebox model. However, prior works that have investigated input transformations for adversarial FR still assumed consistent preprocessing \cite{yang2021towards, zhou2025improving}, and hence the adversarial examples likely overfit to this process.  

Hence, we aim to increase input diversity with respect to the face preprocessing function $A(x)$. We constructed an ensemble loss function that uses $N$ different preprocessing functions $A'(x)$, where $A'(x)$ either applies a crop from a randomly selected face detection model, or a random image resize and downsampling with a randomly selected interpolation method. 

\begin{equation}
\begin{aligned}
  \mathcal {L_\textrm{ensemble}}(x') = \frac{1}{N}\sum^{N}_{i=1}{\mathcal {L}(x', A'_{i})}
\end{aligned}
\label{eq:diverse} 
\end{equation}

For RQ3, we substituted the loss function $\mathcal {L}$ of Equations \ref{eq:lowkey} - \ref{eq:tipim} with our ensemble loss function to verify its effectiveness. By optimising the perturbation over an ensemble of different preprocessing methods, we hypothesise that the adversarial example will be more robust to the preprocessing used in blackbox attack settings. 

\section{Experiments}

\subsection{Experiment Settings}

\textbf{Datasets.} Our experiments are conducted on a subset of the CelebA-HQ dataset \cite{karras2017progressive} containing 3000 images with 300 identities at 1024 x 1024 resolution. This subset was achieved via a stratified sampling process in which 10 images were randomly sampled without replacement for the 300 most frequent identities. The high resolution nature of this dataset was considered crucial in ensuring proper representation of real-world image quality. 

\textbf{Attack Setup.} We consider 11 different preprocessing setups, using seven different face detectors (RQ1) and four different interpolation methods (RQ2). For RQ1 we consistently use area interpolation, and for RQ2 we consistently use MTCNN face detection, to isolate our analysis. We then generate 33 adversarial galleries by running each of the three attacks with each of the 11 preprocessing setting on all images contained within the CelebA-HQ stratified subset. Each gallery contains 3000 adversarial attacked images associated with a unique attack and preprocessing combination. Similarly, for face verification we consider an FR system using each of the 11 different preprocessing setups mentioned. For each of the 300 identities, an image was randomly selected from the 10 images belonging to that identity to serve as a probe image. 

\textbf{Compared Methods.} The ArcFace FR model, as implemented by Yang et al. \cite{yang2020robfr}, is used both for adversarial attack generation and face verification. This model represents the state-of-the-art as indicated by the Face Verification on Labelled Faces in the Wild Benchmark \cite{pwclfw2025}. We consider seven face detection models from the Python DeepFace library \cite{serengil2024lightface}; MTCNN \cite{zhang2016joint}, OpenCV \cite{bradski2000opencv}, Dlib \cite{king2009dlib}, MediaPipe \cite{lugaresi2019mediapipe}, YOLOv8 \cite{redmon2016you}, Centerface \cite{yuanyuan2019centerface}, and RetinaFace \cite{deng2020retinaface}. To reduce the scope of our experiments, we excluded some face detection models that had highly similar computed face regions to each other. We consider four different interpolation methods, through the PyTorch implementation of nearest, bilinear, bicubic, and area, with antialiasing applied where relevant. 

\textbf{Attack Settings.} Attack settings for MIM, LowKey and TIP-IM include maximum perturbation magnitude $\epsilon$, iterations $T$, momentum $\mu$, normalization method, learning rate $\alpha$, and perceptual weighting $\gamma$. Across all the attacks, the normalisation method is set to $L_\infty$ and $\alpha = \frac{1.5 * \epsilon}{T}$, aligning with \cite{yang2020robfr}. Per attack settings are selected to mirror the default settings of each paper \cite{yang2020robfr, cherepanova2021lowkey, yang2021towards}, respectively:

\begin{itemize}
    \item MIM: $\epsilon=8$, $T=100$, $\mu=1.0$.
    \item LowKey:  $\epsilon=8$, $T=50$, $\gamma = 0.05$.
    \item TIP-IM: $\epsilon=12$, $T=50$, $\mu=1.0$. 
\end{itemize} 

For our preprocessing invariant method described in \ref{sec:preprocessing_invariant_method}, we set $N$ to 9, performing 5 different face crops with different face detection models, and 4 different image resizing with different interpolation methods. We sampled each face detection method without replacement from the set of face detection models implemented in DeepFace \cite{serengil2024lightface}. For interpolation, we randomly scale the image by a factor in a range of 0.5 - 2.0, and then sample a random method from nearest, bilinear, bicubic, and area interpolation. 

\begin{figure*}[!htbp]
  \centering
  \includegraphics[width=0.8\textwidth]{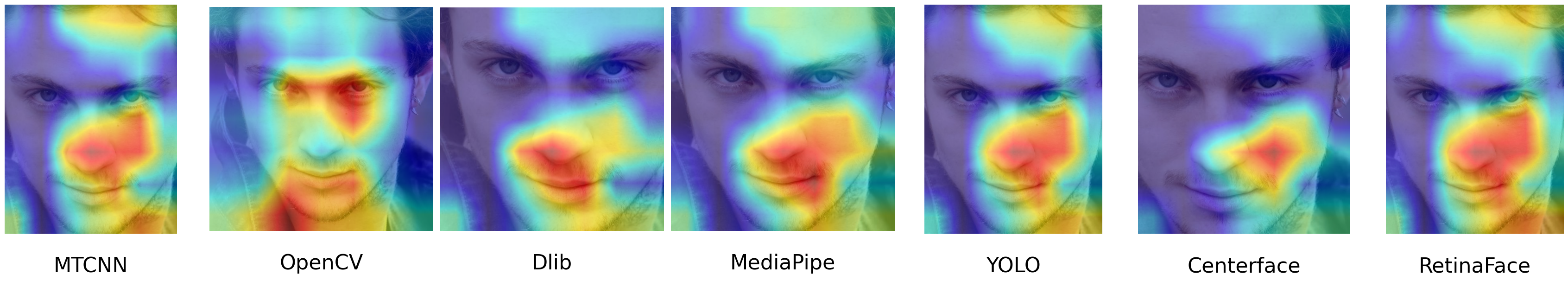}
  \caption{Different crops and their resulting Grad-CAM localisation maps for the ArcFace FR model.}
  \label{fig:Grad-CAM}
\end{figure*}

\textbf{Evaluation Metrics.} We are primarily interested in determining the extent to which the feature distance can be degraded by FR preprocessing. Hence, we define two metrics: \textit{I1}, which is the cosine similarity score of the probe image embeddings against the adversarial example for the same image, and \textit{I9}, which is the average cosine similarity of the probe image embeddings against 9 different images of the same identity. The score ranges between [-1, 1], where a lower score indicates a more effective attack. If the cosine similarity of a probe image and adversarial image is increased, then the adversarial examples are less likely to prevent face matches and fool an FR system. To help measure this, we also use Attack Success Rate (ASR), to see if preprocessing can significantly degrade out-of-the-box attacks against face verification within a FR system. ASR represents the ratio of adversarial examples that successfully evade the FR sytem to the total number of adversarial examples generated. We determine a threshold for ASR based on a FAR@0.05 for our CelebA\_HQ dataset for each individual FR setup. 

\subsection{Effect of Detection Backend (RQ1)}

\textbf{Different face detection models have a significant effect on the strength of adversarial examples against face recognition.} From Table \ref{tab:detection-method-comparison}, we observe that the produced adversarial examples consistently have the strongest impact to facial similarity when the FR system uses the same face detection model as the adversarial attack, as indicated by the diagonal of the table. Inversely, the attack strength is substantially degraded when the FR system uses a different face detection algorithm to the adversarial attack, which we confirm to be significant using a one-way ANOVA test \cite{mcdonald2014handbook} with $p < 0.05$. For instance, the average I1 adversarial feature distance is degraded by up to 197\% (-0.58 → +0.56) for the LowKey attack generated with MTCNN, when applied to a FR system using Dlib face detection. 

Furthermore, this decrease in feature distance of the adversarial examples is significant enough to even affect the ASR for each attack under out-of-the-box settings. Whilst all of the attacks had near perfect attack success rate when using the same face detection model as the target FR system, the ASR was reduced by up to 78\% by preprocessing in blackbox attacks. ASR is also relatively sensitive to the noise strength of the attack. Hence, we would expect the ASR to be degraded much more strongly if the studied attacks incorporated a noise budget, as the average cosine similarity of each attack would be closer to the ASR threshold. 

To understand this result, we use Grad-CAM \cite{selvaraju2017grad} to visualise the localisation maps of the FR model. Figure \ref{fig:Grad-CAM} examines a sample image cropped by different face detectors, each exhibits varying cropping strategies. These differences lead to noticeable variations in the resulting localisation maps, which highlights the most important regions for FR. Notably, crops that differ substantially result in significantly different localisation maps, while similar crops yields more consistent maps  but still exhibit subtle differences. This reveals that the cropping region directly influences FR model’s feature attribution. As a result, perturbations optimised for a specific face crop do not transfer well across detectors, due to the changes in the underlying feature importance. 

\begin{figure}[!htbp]
  \centering
  \includegraphics[width=0.8\columnwidth]{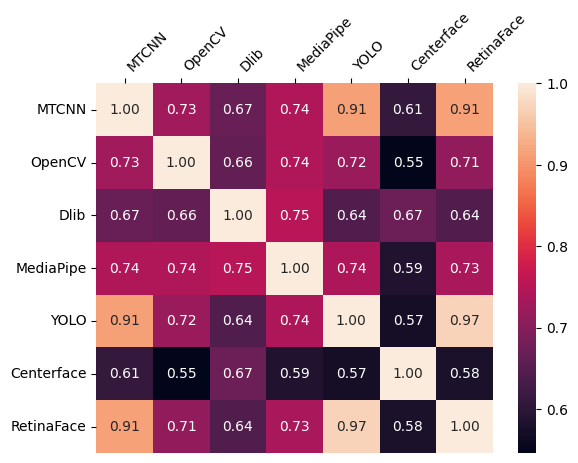}
  \caption{Heatmap of Intersection over Union (IoU) for different face detection models on the CelebA-HQ dataset.}
  \label{fig:heatmap_iou}
\end{figure}

To quantify the differences in face detectors we measure the average Intersection over Union (IoU) between their respective face crop regions across the gallery, as shown in Figure \ref{fig:heatmap_iou}. We find that Centerface and Dlib produce significantly different crops compared to other detectors, reflected by the low IoU score. In contrast, YOLOv8, MTCNN, and RetinaFace show higher mutual overlap and thus more consistent cropping behaviour. We further inspect whether there is a correlation between the percentage change in similarity score and the IoU of the face detection model, by examining the results for when CenterFace is used during adversarial generation in contrast to other face detection models used during evaluation. We observe a weak negative correlation between IoU and attack degradation, with statistical significance confirmed by a coefficient of determination $R^{2}=0.15$ and $p<0.05$ \cite{wright1921correlation}, when examining results for when CenterFace is used during adversarial generation in contrast to other face detection models used during evaluation. These findings support that larger deviations in face crops are associated with greater degradation in attack strength.



\subsection{Effect of Interpolation Method (RQ2)}

\begin{table}[ht]
\centering
\resizebox{\columnwidth}{!}{%
    \begin{tabular}{|c|c|cc|cc|cc|cc|}
    \hline
    \multirow{2}{*}{Interpolation} & \multirow{2}{*}{Attack} & \multicolumn{2}{c|}{Nearest} & \multicolumn{2}{c|}{Bilinear} & \multicolumn{2}{c|}{Bicubic} & \multicolumn{2}{c|}{Area} \\
     &  & I1 & I9 & I1 & I9 & I1 & I9  & I1 & I9  \\
    \hline
    \multirow{3}{*}{Nearest}
     & LowKey & \textbf{-0.55} & \textbf{-0.23} & -0.45 & -0.20 & -0.51 & -0.22 & -0.50 & -0.22 \\
     & MIM    & \textbf{0.87} & \textbf{0.41} & 0.99 & 0.47 & 0.99 & 0.47 & 0.99 & 0.47 \\
     & TIP-IM & 0.29 & 0.21 & 0.27 & 0.20 & \textbf{0.25} & \textbf{0.19} & 0.26 & 0.20 \\
    \hline
    \multirow{3}{*}{Bilinear}
     & LowKey & -0.47 & -0.20 & -0.54 & \textbf{-0.24} & \textbf{-0.56} & \textbf{-0.24} & -0.55 & \textbf{-0.24} \\
     & MIM    & -0.09 & -0.02 & -0.13 & \textbf{-0.04} & \textbf{-0.14} & \textbf{-0.04} & \textbf{-0.14} & \textbf{-0.04} \\
     & TIP-IM & 0.03 & 0.09 & \textbf{-0.01} & 0.08 & \textbf{-0.01} & \textbf{0.07} & \textbf{-0.01} & 0.08 \\
    \hline
    \multirow{3}{*}{Bicubic}
     & LowKey & -0.48 & -0.21 & -0.55 & -0.24 & \textbf{-0.59} & \textbf{-0.26} & -0.57 & -0.25 \\
     & MIM    & -0.11 & -0.04 & -0.14 & -0.05 & \textbf{-0.16} & \textbf{-0.06} & -0.15 & -0.05 \\
     & TIP-IM & 0.03 & 0.09 & -0.01 & 0.08 & \textbf{-0.02} & \textbf{0.07} & -0.01 & \textbf{0.07} \\
    \hline
    \multirow{3}{*}{Area}
     & LowKey & -0.48 & -0.20 & -0.55 & -0.24 & -0.57 & \textbf{-0.25} & \textbf{-0.59} & \textbf{-0.25} \\
     & MIM    & -0.11 & -0.03 & -0.14 & \textbf{-0.05} & \textbf{-0.16} & \textbf{-0.05} & \textbf{-0.16} & \textbf{-0.05} \\
     & TIP-IM & 0.04 & 0.10 & -0.00 & \textbf{0.08} & \textbf{-0.01} & \textbf{0.08} & \textbf{-0.01} & \textbf{0.08} \\
    \hline
    \end{tabular}
}
\vspace{2pt}
\caption{I1 and I9 image similarity scores for different attacks with different interpolation methods. The interpolation method used by the attack and FR system are depicted by rows and columns, respectively. Lower is better for I1/I9 whereas higher is better for ASR. \textbf{Bold values} indicate the best row-wise performance for each individual metric.}
\label{tab:interpolation-method-comparison}
\end{table}

\begin{table*}[ht]
\centering
\renewcommand{\arraystretch}{1.2}
\setlength{\tabcolsep}{4pt}
\scriptsize
\resizebox{1.01\textwidth}{!}{%
    \begin{tabular}{|c|ccc|ccc|ccc|ccc|ccc|ccc|ccc|}
    \hline
    \multirow{2}{*}{Attack} & \multicolumn{3}{c|}{MTCNN} & \multicolumn{3}{c|}{OpenCV} & \multicolumn{3}{c|}{Dlib} & \multicolumn{3}{c|}{MediaPipe} & \multicolumn{3}{c|}{YOLO} & \multicolumn{3}{c|}{Centerface} & \multicolumn{3}{c|}{RetinaFace}\\
     & I1 & I9 & ASR & I1 & I9 & ASR & I1 & I9 & ASR  & I1 & I9 & ASR & I1 & I9 & ASR & I1 & I9 & ASR  & I1 & I9 & ASR \\
    \hline
    LowKey & \textbf{-0.58} & \textbf{-0.25} & \textbf{1.00} & 0.39 & 0.18 & 0.66 & 0.56 & 0.33 & 0.96 & 0.57 & 0.30 & 0.95 & 0.04 & -0.00 & 0.98 & 0.46 & 0.17 & 0.94 & -0.01 & -0.02 & 0.99 \\
    LowKey + Ours & -0.17 & -0.11 & \textbf{1.00} & \textbf{0.16} & \textbf{0.06} & \textbf{0.90} & \textbf{0.27} & \textbf{0.16} & \textbf{1.00} & \textbf{0.29} & \textbf{0.15} & \textbf{1.00} & \textbf{-0.06} & \textbf{-0.06} & \textbf{1.00} & \textbf{0.20} & \textbf{0.07} & \textbf{1.00} & \textbf{-0.07} & \textbf{-0.06} & \textbf{1.00} \\
    \hline
    MIM & \textbf{-0.16} & \textbf{-0.05} & \textbf{1.00} & 0.36 & 0.17 & 0.69 & 0.52 & 0.30 & 0.97 & 0.54 & 0.28 & 0.96 & 0.16 & 0.07 & 0.98 & 0.44 & 0.17 & 0.95 & 0.13 & 0.06 & 0.98 \\
    MIM + Ours & 0.00 & 0.01 & \textbf{1.00} & \textbf{0.20} & \textbf{0.09} & \textbf{0.89} & \textbf{0.32} & \textbf{0.21} & \textbf{1.00} & \textbf{0.33} & \textbf{0.19} & \textbf{1.00} & \textbf{0.09} & \textbf{0.04} & \textbf{1.00} & \textbf{0.27} & \textbf{0.12} & \textbf{0.98} & \textbf{0.08} & \textbf{0.04} & \textbf{1.00} \\
    \hline
    TIP-IM & -0.01 & 0.08 & \textbf{0.99} & 0.22 & 0.12 & 0.84 & 0.46 & 0.30 & \textbf{0.99} & 0.48 & 0.29 & \textbf{0.98} & 0.11 & 0.10 & \textbf{0.99} & 0.38 & 0.17 & 0.96 & 0.09 & 0.09 & 0.99 \\
    TIP-IM + Ours & \textbf{-0.02} & \textbf{0.07} & \textbf{0.99} & \textbf{0.16} & \textbf{0.09} & \textbf{0.90} & \textbf{0.41} & \textbf{0.28} & \textbf{0.99} & \textbf{0.43} & \textbf{0.27} & \textbf{0.98} & \textbf{0.06} & \textbf{0.08} & \textbf{0.99} & \textbf{0.33} & \textbf{0.15} & \textbf{0.97} & \textbf{0.05} & \textbf{0.08} & \textbf{1.00} \\
    \hline
    \end{tabular}
}
\vspace{2pt}
\caption{I1/I9 image similarity scores and Attack Success Rate (ASR) for out of the box attacks using MTCNN face detection and area interpolation, in-comparison to our preprocessing-invariant method. \textbf{Bolded} values indicate the best columnwise metrics for each FR face detection backend for each attack.}
\label{tab:rq3-method-comparison}
\end{table*}

\begin{figure}[!htbp]
  \centering
  \includegraphics[width=\columnwidth]{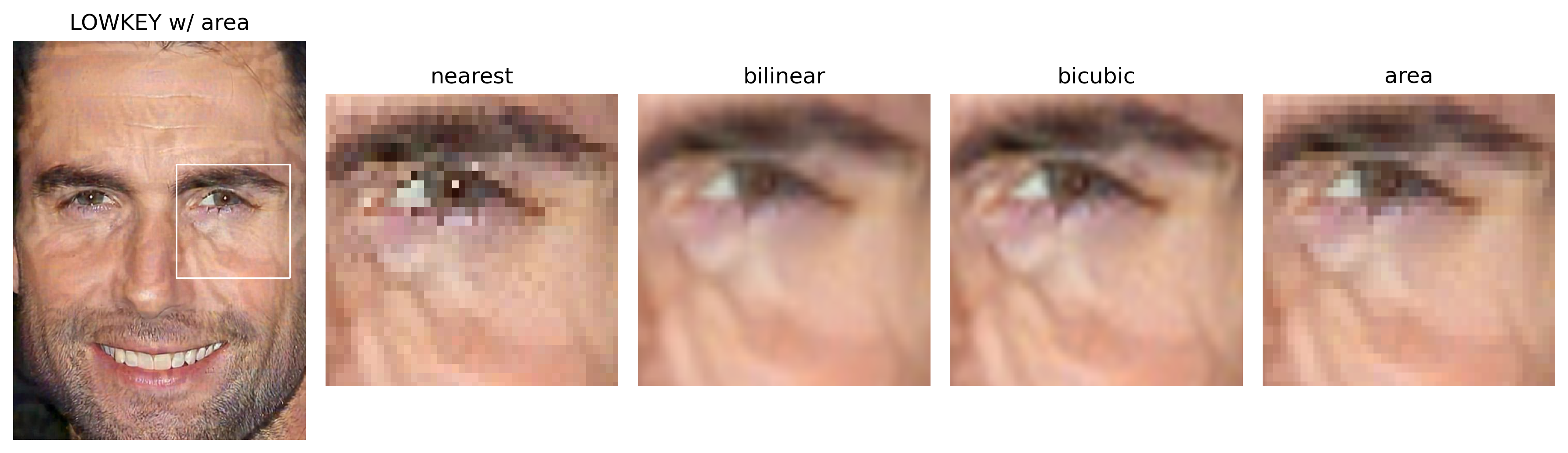}
  \caption{Visual comparison of information loss for different interpolation methods used when downsampling a LowKey adversarial example.}
  \label{fig:interpolation_FR_comparison}
\end{figure}

\textbf{Interpolation does not have a significant effect on generalisation of adversarial examples for face recognition.} From Table \ref{tab:interpolation-method-comparison} we observe that there is minimal difference in noise vectors produced by attacks using different interpolation methods, as indicated through the minimal differences for I1 and I9 metrics within each attack setup. Consequently, ASR is unaffected by the change in interpolation. 



To better understand this result, we performed visual inspection of adversarial examples after being downsampled by different interpolation methods; as shown in Figure \ref{fig:interpolation_FR_comparison}. We observed that whilst interpolation alters the quality of the image, the noise pattern remains relatively unaffected, which is reflected by the insignificant change in the attack strength. However, our results indicate that \textit{bicubic} interpolation preserved information slightly better, as this method produced the lowest image similarity on average. 

\subsection{Preprocessing Invariant Method (RQ3)}
\textbf{Preprocessing dependent image transformations significantly improve adversarial generalisation against preprocessing in blackbox attacks.} Table \ref{tab:rq3-method-comparison} displays the comparison of the performance of out-of-the-box adversarial attacks in comparison to the same adversarial attacks with our added input transformations. We observe that our preprocessing invariant adversarial method universally improved the transferability of the adversarial attacks against different face detection models, through a greater feature distance produced by the adversarial images. This performance improvement was also significant enough to affect the ASR of each attack under each setup. Whilst our method performs worse under a whitebox setup in which the preprocessing matches the target FR system, as shown by the results for MTCNN in Table \ref{tab:rq3-method-comparison}, this decrease in performance was not significant enough to degrade the ASR however. Notably, our method also transfers better than the original TIP-IM attack, which applies generic affine transformations to adversarial perturbations at each iteration. Our method also produces similar perceptual similarity to the original attacks, so it works at a similar noise budget. The average Peak Signal to Noise Ratio (PSNR) of the original attacks compared to our method was $12.54 \rightarrow 12.50$. 

\subsection{Effect of Adversarial Examples on Face Detection}

\textbf{Face preprocessing can even impact the effectiveness of an adversarial attack in a whitebox setting.} To remove potential confounding variables in our results, we did not recalculate face regions for adversarial examples after the adversarial perturbation is applied, so that we could analyse the effect of the face crop region in isolation. However, in practice an FR system would need to recalculate any face regions using its own preprocessing pipeline. We observe that the perturbations introduced by an adversarial attack have an unintended consequence on the face detection model and cause a subtle shift in the detected face region, in comparison to the original image. To investigate this further, we recalculate the face region using MTCNN for all gallery images produced using MTCNN face detection, to investigate the impact the noise vector can have. 

\begin{table}[ht]
\centering
    \begin{tabular}{|c|c|cc|}
    \hline
    Attack & IoU & \multicolumn{2}{c|}{I1} \\
    \cline{3-4}
    & & Original & Adversarial\\
    \hline
    LowKey & 0.94 & -0.58 & -0.30 \\
    MIM    & 0.93 & -0.16 & 0.00 \\

    TIP-IM & 0.93 & -0.01 & 0.00 \\
    \hline
    \end{tabular}
\vspace{2pt}
\caption{IoU and I1 scores for original and adversarial face regions detected using the same face detection model, across different attacks. I1 scores calculated with area interpolation and MTCNN face detector.}
\label{tab:face-crop-shift-analysis}
\end{table}

Table \ref{tab:face-crop-shift-analysis} indicates that even when using the same original image and the same face detection model, the adversarial examples only have an average IoU of ~0.93. Perturbations introduced during the attack have an unintended effect on the image features for the face detector, causing the detected face region to shift. This translates to a significant reduction in attack effectiveness in MIM and LowKey as demonstrated by an increase in cosine similarity of 0.16 ($-0.16 \rightarrow 0$) and 0.28 ($-0.58 \rightarrow -0.30$), respectively, when comparing evaluation using a consistent face region for both images (original) and evaluation with the face region calculated individually. This reduction in attack strength points toward overfitting of perturbations produced during the attack to the spatial region identified in the original image. These results demonstrate the significance of face preprocessing, as it can heavily degrade the effectiveness of the adversarial perturbation even in a whitebox setting, where the same face preprocessing techniques and models are used. 

The I1 score for the TIP-IM attack was much less heavily degraded despite a similar difference in IoU. As with RQ1 and RQ3, this potentially highlights the effectiveness of input transformations for improving transferability. 

\section{Conclusion \& Future Work}

We studied the impact of blackbox preprocessing against adversarial examples for FR systems. Our extensive experiments demonstrated that facial image preprocessing plays a significant role in adversarial attacks and can rapidly degrade adversarial image embedding distances. We found that input transformations are an effective solution against this problem however, improving the adversarial transferability. 

In future, we intend to investigate the impact of additional facial preprocessing steps, such as normalisation and alignment to obtain a more complete understanding of these impacts. We additionally aim to consider this problem in combination with blackbox FR models, to provide end-to-end investigation of adversarial generalisation.

\section*{Acknowledgment}
We would like to acknowledge Dmitri Kamenetsky, Victor Stamatescu, and the Defence Science Technology Group for their support and review of this work.

\bibliographystyle{IEEEtran}
\bibliography{bibfile}

\end{document}